\documentclass[a4paper,final,12pt,onecolumn]{IEEEtran}
\usepackage{setspace}
   \usepackage[dvips]{graphicx}
   \graphicspath{{../eps/}}
   \DeclareGraphicsExtensions{.pdf,.jpeg,.png,.eps}

\usepackage[cmex10]{amsmath}
\usepackage[tight,footnotesize]{subfigure}

\begin{document}
\title{memristive fuzzy edge detector}
\author{Farnood~Merrikh-Bayat,
        Saeed Bagheri Shouraki
        \thanks{F. Merrikh-Bayat and S. Bagheri Shouraki are with the Department
of Electrical Engineering, Sharif University of Technology, Tehran, Iran, e-mails: f\_merrikhbayat@ee.sharif.edu and bagheri-s@sharif.edu.}}

%
\maketitle

\begin{abstract}
Fuzzy inference systems always suffer from the lack of efficient
structures or platforms for their hardware implementation. In this
paper, we tried to overcome this problem by proposing new method for
the implementation of those fuzzy inference systems which use fuzzy
rule base to make inference. To achieve this goal, we have designed
a multi-layer neuro-fuzzy computing system based on the memristor
crossbar structure by introducing some new concepts like fuzzy
minterms. Although many applications can be realized through the use
of our proposed system, in this study we show how the fuzzy XOR
function can be constructed and how it can be used to extract edges
from grayscale images. Our memristive fuzzy edge detector
(implemented in analog form) compared with other common edge
detectors has this advantage that it can extract edges of any given
image all at once in real-time.
\end{abstract}

%

\section{introduction}

In the past decades, the integrated circuits (IC) industry has
successfully followed Moore's Law \cite{moore}. However,
Complementary Metal Oxide Semiconductors (CMOS) scaling is
approaching a physical and economical limit. To effectively extend
Moore's law, in addition to pushing the limit of lithography for
smallest possible devices, there is a great need for more powerful
devices, disruptive fabrication technologies, alternative computer
architecture and advanced materials, etc.

One possible way to extend this law beyond the limits of transistor
scaling is to obtain the equivalent circuit functionality using an
alternative computing scheme. Nowadays, most of the currently
working computing systems like Digital Signal Processors (DSPs) and
Field Programmable Gate Arrays (FPGAs) are constructed based on two
basic concepts; they use digital logic to perform computing or
decision making tasks and they work in discrete form. The former
results in a separation of memory and computing units \cite{Moneta}
and inefficient computation while the later one leads to slow and
high area-consuming systems. Such computing paradigms usually suffer
from the constant need of establishing a trade-off between
flexibility and performance. They also introduce limited numerical
precision both in the input signals and the algorithmic quantities.
For example, input signals are usually quantized to limited numeric
precision in A/D converters. In addition, the arithmetic operations
are carried out with limited computational precision and the results
are rounded or truncated to a specific limited precision
\cite{cioffi,haykin,Treichler}.

In recent years, particularly after the publication of the paper
\cite{williams} in 1 May 2008, we have seen considerable scientific
and technological progress in the field of memristive computing
systems. This great interest in these systems is mostly due to their
potential in overcoming most of the aforementioned challenges in
front of today's digital systems which has nominated them as an
alternative computing scheme. For example, it has been demonstrated
that these systems can be constructed much denser and faster through
the use of nano-crossbar technology and they consume much less
energy than their counterparts \cite{snider2}. However, almost all
of these systems was again constructed based on the concepts of
traditional digital logic.

Recently, we showed that it is possible to design a memristive soft
computing system \cite{farnoodefficient} with learning capabilities
which uses fuzzy logic instead of digital logic to do its
computations. In addition to be build on fuzzy logic's concepts, it
was implemented in analog form and therefore it was very fast and
completely consistent with the analog nature of memristor. Moreover,
our proposed neuro-fuzzy computing system had this advantage that in
its hierarchical structure, memory units were assimilated with
computational units like what we have in human brain. Now, in this
paper, we will show another way to implement fuzzy inference systems
that use fuzzy rule base to make inference by introducing new
concepts like {\it fuzzy minterms}. Although this analog multi-layer
neuro-fuzzy system which is somehow inspired from our earlier work
can be used in so many image processing tasks, here we only
concentrate on the application of edge detection from grayscale
images. For this purpose, at first we will describe the fuzzy XOR
function by fuzzy rule base. Then, this fuzzy function will be
constructed through our multi-layer neuro-fuzzy system and then will
be applied to consecutive pixels of the input grayscale image to
extract edges from it. As simulation results indicate, our proposed
method extracts much sharper and meaningful edges compared with
traditional edge detecting algorithms even in noisy environment.
However, the main benefit of our fuzzy edge detector is for its
efficient hardware implementation in analog form. Actually, this is
because of this advantage that this structure can detect all
horizontal and vertical edges in grayscale images simultaneously in
real-time. Finally, it should be noted that since our computing
system is constructed by the use of memristor crossbars, it can be
simply reconfigured even during its working time by the
reprogramming of memristors in crossbars.

This paper is organized as follows. The working procedure of
memristor crossbar structures and their application in the hardware
implementation of artificial neural networks are described in
Section \ref{memristor}. The process of constructing binary XOR
function by using the network proposed by McCulloch-Pitts and the
problem of thresholding in these networks are demonstrated in
Section \ref{sec22}. Section \ref{fuzzyxorgate} is devoted to the
explanation of our proposed multi-layer neuro-fuzzy computing system
designed for the implementation of those fuzzy inference systems
which use fuzzy rule base to make inference. Application of the
constructed fuzzy XOR function in detecting edges from grayscale
images is presented in Section \ref{edge}. Eventually, some
experimental results are presented in Section \ref{simulation},
before conclusions in Section \ref{conclusion}.

\section{Memristor crossbars}
\label{memristor} After the first experimental realization of the
fourth fundamental circuit element, {\it i.e.} memristor, in its
passive form \cite{williams}, whose existence was previously
predicted in 1971 by Leon Chua \cite{Chua}, many researches are
seeking its applications in variety of fields such as neuroscience,
neural networks and artificial intelligence. It has become clear
that this passive element can have many potential applications such
as creation of analog neural network and emulation of human learning
\cite{pershin}, building programmable analog circuits
\cite{perShin22,farelsevmemarith}, constructing hardware for soft
computing tools \cite{farIEEE}, implementing digital circuits
\cite{kuekes} and in the field of signal processing \cite{Mouttet1,
Mouttet2}.

From the mathematical model of memristor (for example the one
reported by HP \cite{williams}), it can be concluded that passing
current from memristor in one direction will increase the
memristance of this passive element while changing the direction of
the applied current will decrease its memristance. In addition,
passing current in one direction for longer period of time will
change the memristance of the memristor more. Moreover, by stopping
the current from passing through the memristor, memristance of the
memristor will not change anymore. As a result, memristor is nothing
else than the analog variable resistor where its resistance can be
properly adjusted by changing the direction and duration of the
applied voltage. Therefore, memristor can be used as a simple
storage device in which analog values can be stored as a memristance
instead of voltage or charge.

\begin{figure}[!t]
\centering 
{
\includegraphics[width=2.5in,height=1.4in]{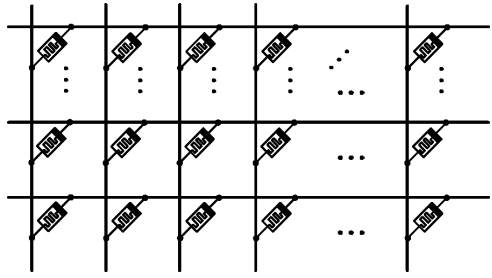}}
\caption{A typical memristor crossbar.}
\label{fig2} 
\end{figure}

A crossbar array basically consists of two sets of conductive
parallel wires intersecting each other perpendicularly. The region
where a wire in one set crosses over a wire in the other set is
called a crosspoint (or junction). Crosspoints are usually separated
by a thin film material which its properties such as its resistance
can be changed for example by controlling the voltage applied to it.
One of such devices is memristor which is used in our proposed
circuits in this paper. Figure \ref{fig2} shows a typical memristor
crossbar structure. In this circuit, memristors which are formed at
crosspoints are depicted explicitly to have better visibility. In
this crossbar, memristance of any memristor can simply be changed by
applying suitable voltages to those wires that memristor is
fabricated between them. For example, consider the memristor located
at coordinate (1, 1) (crossing point of the first horizontal and the
first vertical wires) of the crossbar. Memristance of this memristor
can be decreased by applying a positive voltage to the first
vertical wire while grounding the first horizontal one (or
connecting it to a negative voltage). Dropping positive voltage
across one memristor will cause the current to pass through it and
consequently, memristance of this passive element will be decreased.
In a similar way, memristance of this memristor can be increased by
reversing the polarity of the applied voltage. As stated before,
application of higher voltages for longer period of time will change
the memristance of the memristor more. This means that the
memristance of any memristor in the crossbar can be adjusted to any
predetermined value by the application of suitable voltages to
specific row and column of the crossbar.

To summarize, memristor crossbar is a 2-dimensional grid that analog
values can be stored in its crosspoints through the memristance of
the memristors. Consequently, it seems that the memristor crossbar
is a perfect structure to construct and store 2-dimensional weight
matrix of neural networks \cite{afifi,snider} as used in these paper
as well.

\section{Using artificial neural networks to construct a binary XOR function and the problem of thresholding in these networks}
\label{sec22}

\begin{figure}[!t]
\centering \subfigure[]{
\label{fig1a} 
\includegraphics[width=3in,height=1.1in]{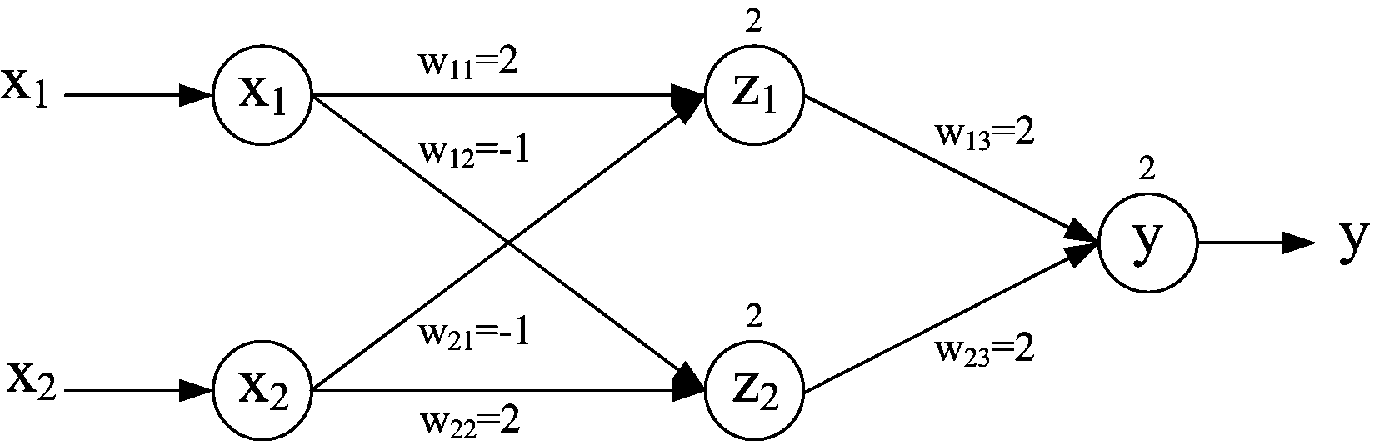}}
 \subfigure[]{
\label{fig1b} 
\includegraphics[width=2.3in,height=2.3in]{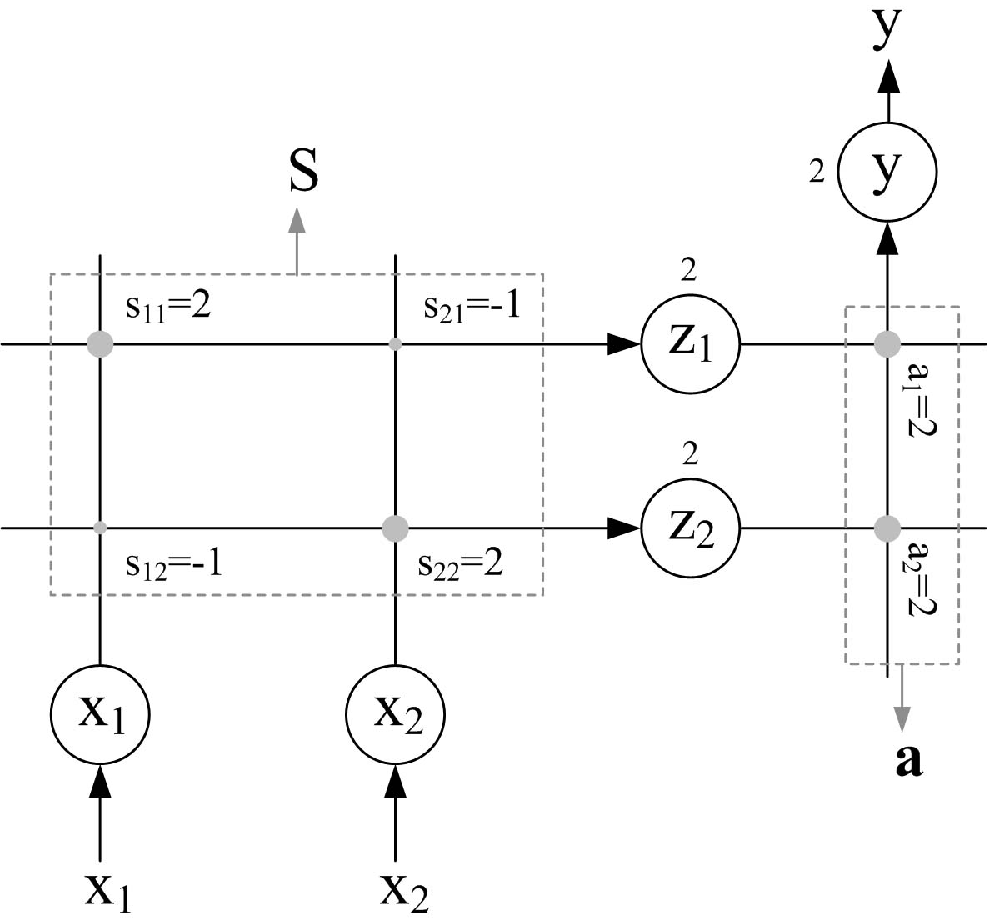}}
\subfigure[]{
\label{fig1c} 
\includegraphics[width=2.9in,height=2.5in]{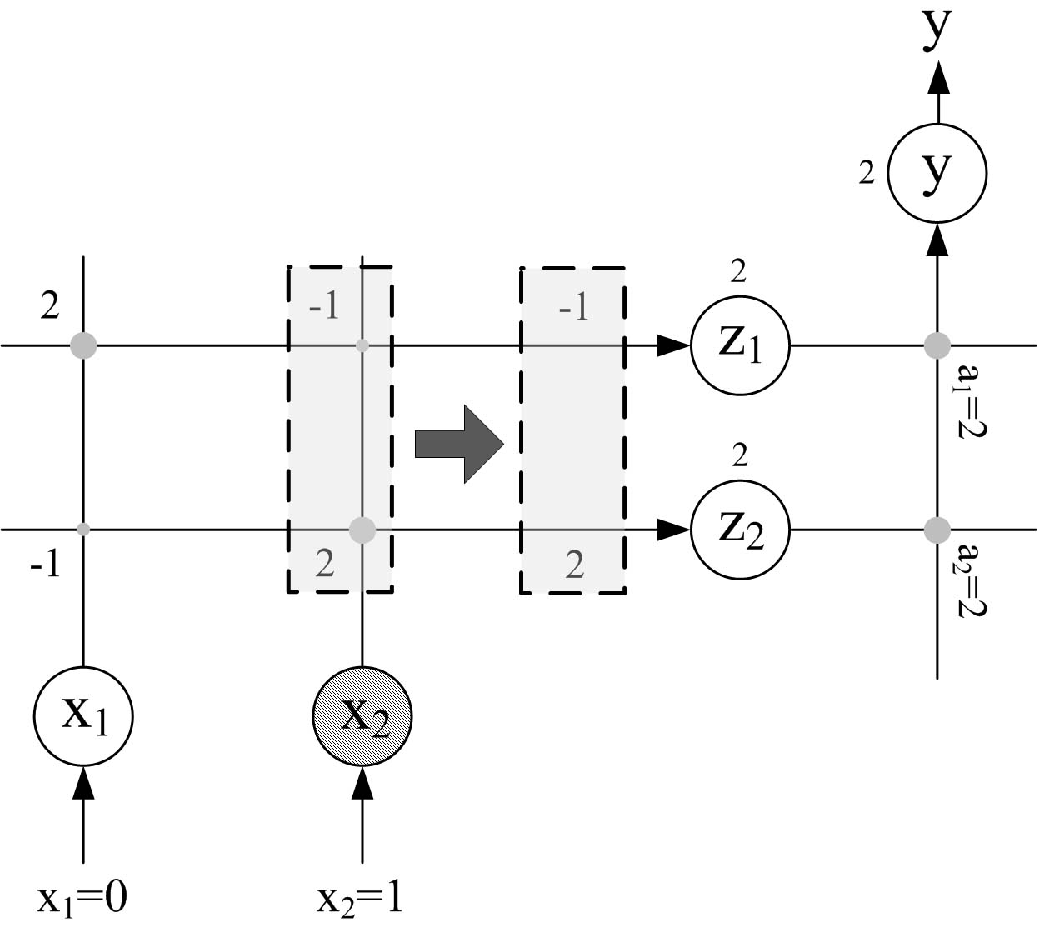}}
 \caption{(a) The artificial neural network proposed by McCulloch-Pitts for the implementation of logical XOR function.
 (b) The same network of Fig. \ref{fig1a} redrawn by the usage of nano-crossbar structures. (c) This figure shows that
 the connection weight matrix can be treated as a fuzzy relation. In this case, activation of one concept at the input
 layer will select one column of the fuzzy relation as an output.}
\label{fig1} 
\end{figure}

 Since our goal is to propose a simple structure for
the fuzzy XOR function, it will be very useful to see how binary XOR
gate is implemented in primary artificial neural networks. For this
purpose, consider one of the simplest networks proposed by
McCulloch-Pitts in 1943 \cite{McCulloch,fasset} for the
implementation of logical XOR gate which is depicted in Fig.
\ref{fig1a} for convenience. In this figure, each neuron with binary
activation receives a number of inputs (either from original data or
from the output of other neurons in the networks). Each of these
inputs comes via a connection that has a strength (or weight); these
weights correspond to synaptic efficiency in a biological neuron. In
each neuron, the weighted sum of inputs is formed and then a simple
hard thresholding function is applied to the sum to produce the
output result of the neuron. In the network of Fig. \ref{fig1a}, all
neurons except neurons of the input layer have a threshold value of
2 (as written near to them on the figure) which means that when
their total input (weighted sum of inputs) becomes more than 2,
their output will be set to logic 1. Otherwise, their response will
become equal to logic 0. As a result, this network performs the
logical XOR function on two binary inputs, {\it i.e.} $x_1$ and
$x_2$, and creates binary output $y$.

Note that in the network shown in Fig. \ref{fig1a}, each neuron with
its corresponding connection weights performs a simple logic
function. For example, in the hidden layer, neuron $z_1$ with
connection weights of $w_{11}$ and $w_{21}$ creates logic function
$x_1\overline{x_2}$ ($x_1$ AND NOT $x_2$) while the neuron $z_2$
with connection weights of $w_{12}$ and $w_{22}$ builds logic
function $\overline{x_1}x_2$ ($x_2$ AND NOT $x_1$). Also, in the
output layer, neuron $y$ with connection weights of $w_{13}$ and
$w_{23}$ creates the logical OR function on the outputs of neurons
$z_1$ and $z_2$. Therefore, the overall network of Fig. \ref{fig1a}
performs the following logic function on its inputs $x_1$ and $x_2$:
\begin{equation}\label{eq1}
y=x_1\ XOR\ x_2=x_1\oplus
x_2=x_1\overline{x_2}+x_2\overline{x_1}=\left(x_1 \ AND\  NOT \
x_2\right)\ OR\ \left(x_2 \ AND\  NOT\  x_1\right)
\end{equation}
or equivalently it can be expressed as:
\begin{equation}\label{eq2}
y=\left(x_1\wedge \neg x_2\right)\vee\left(x_2\wedge \neg x_1\right)
\end{equation}

Based on these equations, it can be said that the neural network of
Fig. \ref{fig1a} implements the logical XOR function in the
\textit{sum of products} form. However, this network compared with
traditional digital circuits has this advantage that by some
modification, it can have learning capability.

Note that we have written Eq. \eqref{eq1} in the form of  Eq.
\eqref{eq2} to show the probable similarities between tasks which
are done in traditional neural networks with those tasks that fuzzy
logic usually performs. For this purpose, we can interpret the
working procedure of the network of Fig. \ref{fig1a} in another way.
Figure \ref{fig1b} shows the same network but in a different form.
In this figure, connection weights between layers are implemented
through the nano-crossbar structures. A complete description of the
procedure of the construction of synaptic weights through the usage
of memristor crossbar is provided in \cite{farIEEE} and
\cite{farnoodefficient}. Let's denote those connection weights which
are located between the input and the hidden layer by a $2\times 2$
matrix $S$ and those connection weights which are located between
the hidden and the output layer by a $2\times 1$ vector
$\mathbf{a}$. In addition, let's assume that each neuron in the
network represents one numerical or linguistic concept. For example,
in the network of Fig. \ref{fig1b}, the output neuron may represent
concepts like ``red'', ``small'', ``age=23'' and etc. In this case,
it can be said that when the output of one neuron becomes active
(logic 1), it shows that its corresponding concept is happened.
Therefore, by this way we have changed the meaning of the output of
neurons. In the other words, we have assumed that the output of each
neuron at any time shows our confidence degree about the occurrence
of the concept assigned to that neuron. Consequently, by considering
the working procedure of the network of Fig. \ref{fig1b} in this
way, it becomes clear that signals which are propagating in the
network will be in the form of confidence or membership degrees. In
this case, the connection weight matrix $S$ and the connection
weight vector $\mathbf{a}$ will somehow have the role of fuzzy
relations in fuzzy inference systems \cite{farnoodefficient}. To
clarify this matter further, consider the following example. Assume
that the inputs of the McCulloch-Pitts neural network of Fig.
\ref{fig1b} are $x_1=0$ and $x_2=1$ where for this configuration of
inputs, the output of the system should be $y=1$ (remember that this
network implements the logical XOR function). Figure \ref{fig1c}
shows the first step in which inputs are applied to the network.
Setting $x_2$ to logic 1 means that the confidence degree about the
occurrence of that concept which is assigned to neuron $x_2$ is 1 or
equivalently 100 percent. On the other hand, by setting $x_1$ to
logic 0 we say that we are absolutely sure that the concept assigned
to neuron $x_1$ has not happened or our confidence degree about the
occurrence of this concept is zero. Based on the current output
values of neurons $x_1$ and $x_2$, input of neurons $z_1$ and $z_2$
can be computed which will be equal to -1 and 2 respectively as
shown in Fig. \ref{fig1c} as well. Note that if we put these two
values in a vector like $[-1 \quad 2]^T$, it will become equal to
the second column of matrix $S$ which can be also obtained through
the simple vector to matrix multiplication as follows:
\begin{eqnarray}\label{eq3}
[-1\quad 2]^T=S\times [0\quad 1]^T=\left(
                                     \begin{array}{cc}
                                       2 & -1 \\
                                       -1 & 2 \\
                                     \end{array}
                                   \right)\times[0\quad 1]^T
\end{eqnarray}
where in this equation, $[0 \quad 1]^T$ is the outputs of neurons
$x_1$ and $x_2$ in a vector form. Now, if we assume that matrix $S$
is a fuzzy relation connecting vector variables $\mathbf{x}=[x_1\
x_2]$ and $\mathbf{z}=[z_1\ z_2]$, then the vector to matrix
multiplication of Eq. \eqref{eq3} will be equal to the sum-product
fuzzy inference method \cite{kosko}. Again, inputs of neurons $z_1$
and $z_2$ will have the role of confidence degrees. For instance,
when in this example the input of neuron $z_2$ becomes 2, it means
that the confidence degree that the occurrence of the concept
assigned to neuron $x_1$ may result in the occurrence of the concept
assigned to neuron $z_1$ is high. On the other hand, when the input
of neuron $z_2$ is -1, it means that those concepts which are
assigned to neurons $x_2$ and $z_2$ are independent from each other
and we do not have belief that the firing of neuron $x_1$ results in
the firing of neuron $z_1$. In the other words, firing of neuron
$x_1$ activates neuron $z_2$ with higher confidence degrees, {\it
i.e.} 2, compared with other neurons which are located in the same
layer.

In the next step, each neuron puts a threshold on its input. There
are two main reasons behind this thresholding task in traditional
neural networks. First, by this way it is possible to force output
of neurons to be bounded between two specific values (in the special
case of the hard thresholding of this example, output of neurons
will be either 0 or 1). Second, thresholding of the outputs of
neurons will eliminate those neurons which have low output value and
therefore inserts nonlinearities to the system. As a result, these
neurons cannot affect the rest of the network while the effect of
those neurons which their outputs are above the threshold value will
be strengthened. However, this thresholding task has a simple
problem: what should be the threshold value? In traditional neural
network it is common to determine the threshold value through the
training process. Now, let's see how this thresholding task can be
modified in networks that deal with confidence degrees (as explained
before) without degrading the performance of the network
significantly. In fuzzy logic, it is well-known that confidence or
membership degrees are always non-negative and there is no necessity
for the height of fuzzy sets or numbers to be equal to one.
Therefore, in networks which are working with confidence degrees,
instead of common thresholding functions such as a binary or bipolar
sigmoid function, other functions can be used. However, these
functions should have the aforementioned property: output of neurons
with higher output value should be strengthened more than those
neurons which have lower output value. For example, one of such
functions is $f(x)=x^n$ where $n$ can be any real number greater
than 1. It is clear that using these functions instead of common
threshold functions can have this benefit that they do not have any
variable parameter to be determined like the threshold value.

Based on the explanations provided in this section, we will modify
the network of Fig. \ref{fig1b} to create a new structure as the
hardware implementation of fuzzy XOR function which is capable of
working with signals of a kind of confidence degree.

\section{fuzzy Exclusive OR (XOR) function and its memristor crossbar-based hardware implementation}
\label{fuzzyxorgate}

In this section, we want to present a new way to build fuzzy logic
functions specially the fuzzy Exclusive OR (XOR) gate. Actually, we
want to show how a fuzzy version of the McCulloch-Pitts network
shown in Fig. \ref{fig1a} can be constructed. First of all, note
that the fuzzy XOR function between input variables $x_1$ and $x_2$
can be expressed through the following fuzzy rule base:

IF $\quad$ $x_1$ is \textit{small} $\quad$ AND $\quad$ $x_2$ is
\textit{small} $\quad$ THEN $\quad$ $y$ is \textit{small}

IF $\quad$ $x_1$ is \textit{small} $\quad$ AND $\quad$ $x_2$ is
\textit{big} $\quad\ \ \ $ THEN $\quad$ $y$ is \textit{big}

IF $\quad$ $x_1$ is \textit{big} $\quad\ \ $ AND $\quad$ $x_2$ is
\textit{small} $\quad$ THEN $\quad$ $y$ is \textit{big}

IF $\quad$ $x_1$ is \textit{big} $\quad\ \ $ AND $\quad$ $x_2$ is
\textit{big} $\quad\ \ $ THEN $\quad\ $ $y$ is \textit{small}

It is clear that any other fuzzy system which is expressed based on
fuzzy rule base can be constructed in a similar way by utilizing our
proposed method as described below.

\begin{figure}[!t]
\centering
\includegraphics[width=6.5in,height=5in]{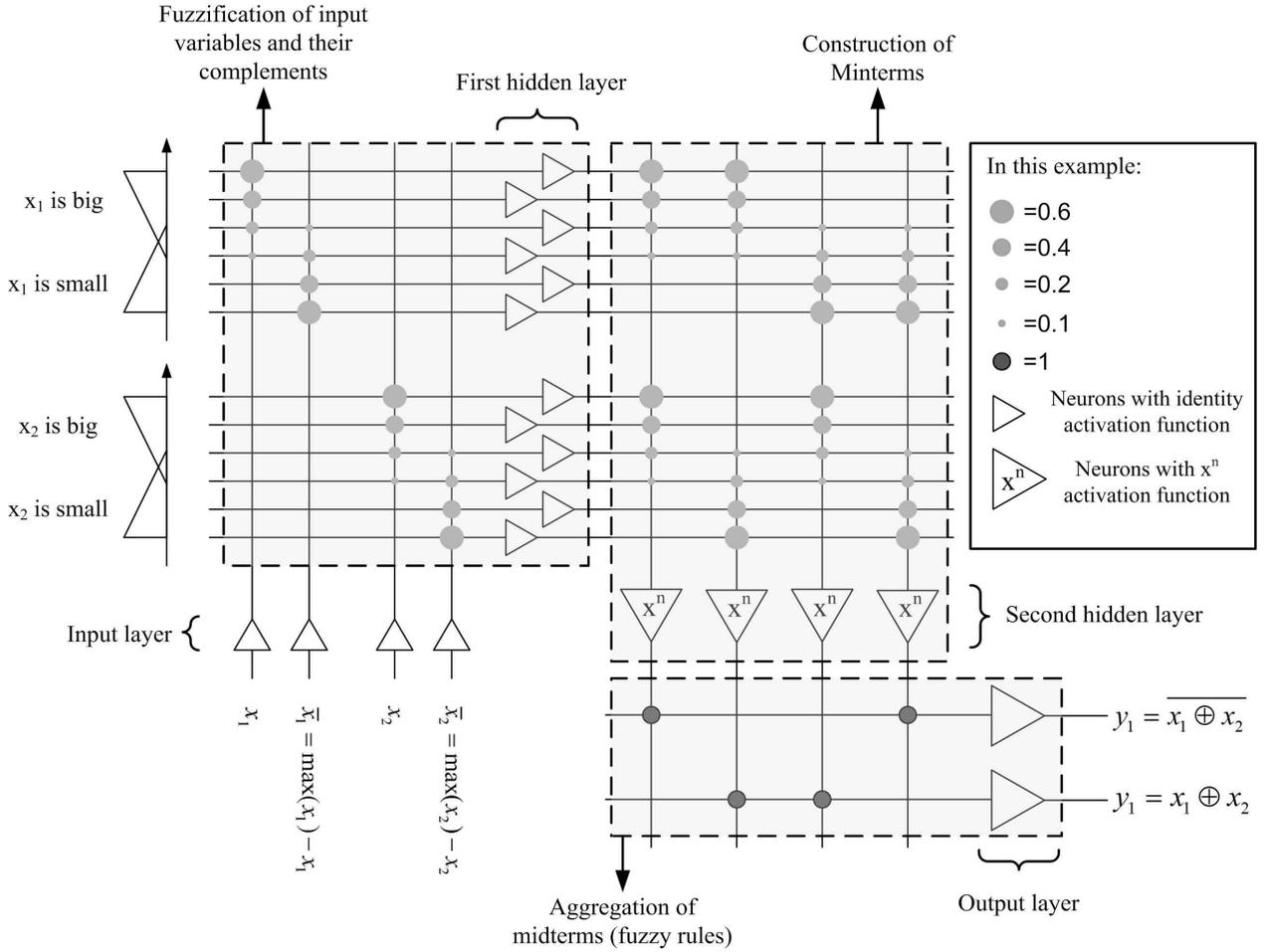}
\caption{Our proposed structure for the hardware implementation of
the fuzzy XOR function.}
\label{fuzzyXOR} 
\end{figure}

Based on the explanations provided in Section \ref{sec22}, we have
proposed a memristor crossbar-based hardware for the fuzzy XOR
function. This hardware is shown in Fig. \ref{fuzzyXOR}. The circuit
of Fig. \ref{fuzzyXOR} consists of three different parts which are
the fuzzification, the fuzzy minterm creating and the aggregation
units. In the next three subsections, the working procedure of each
of these parts is explained.

\subsection{The fuzzification unit}

Since inputs and outputs of most of currently working systems are
crisp, we need to find a way to convert them to their corresponding
fuzzy numbers before using them in our structure. Consequently, the
first part of the circuit of Fig. \ref{fuzzyXOR} (from the input
layer to the first hidden layer specified by a gray dashed
rectangle) is intended for this purpose.

In the antecedent parts of the fuzzy rule base describing the
working procedure of the fuzzy XOR function, two different concepts
or fuzzy sets are considered for each of the input variables which
are ``small'' and ``big''. At any time, based on the observed values
for input variables, some of these concepts become active with
different strengths. For example, when input variable $x_1$ has its
maximum value, its corresponding ``small'' and ``big'' concepts
should become active with minimum and maximum possible strengths or
confidence degrees respectively. For this reason, we have considered
two distinct input terminals for each of the input variables in the
fuzzification unit of the circuit of Fig. \ref{fuzzyXOR}; one for
concept ``$x_i$ is big'' and one for concept ``$x_i$ is small''
where in this structure $i$ can be either 1 or 2. In this case, each
of these input terminals will specify one individual concept and
those values that we apply to them will somehow determine our
confidence degrees about the occurrence of these assigned concepts
for given input data. Actually, in this way, we treat input data as
confidence degrees and not as a meaningless value of crisp
variables. Therefore, in the circuit of Fig. \ref{fuzzyXOR}, input
and output values of the system are of a kind of membership or
confidence degrees and it is the combination of these values and
those concepts which are assigned to input and output neurons that
creates fuzzy numbers. For example in the application of edge
detection, the first and the second input neurons of the system of
Fig. \ref{fuzzyXOR} will represent concepts ``brightness of the
first input pixel'' and ``darkness of the first input pixel''
respectively. In this case, when intensity values of pixels are
applied to these neurons as an input, they will be interpreted as
the confidence degree of those aforementioned concepts. Therefore,
henceforward when the intensity value of one pixel is 255, it will
not show the brightness of the pixel anymore but it will demonstrate
that our confidence degree about the brightness of this pixel is
maximum or 255. On the other hand, applying 255 to the input neuron
which is the representative of the concept ``darkness on the first
input pixel'' means that our confidence degree about the darkness of
this pixel is maximum and no other pixel in the image can be darker
than this one. Therefore, it should become clear that in our
network, applied input and generated output values do not have any
meaning by themselves alone and they only specify the strength of
the activation of those concepts which are assigned to neurons.

Since the two fuzzy sets defined on the universe of discourses of
input variables, {\it i.e.} ``small'' and ``big'', are dependent on
each other (because when one variable is not ``big'' it will be
``small''), values which are applied to their corresponding input
terminals should be dependent as well. For this purpose, in the
structure of Fig. \ref{fuzzyXOR}, we apply the current value of
variable $x_i$ to the input terminal representing concept ``$x_i$ is
big'' and its complement which is defined as $x_i^{max}-x_i$ where
$x_i^{max}$ is the maximum value that variable $x_i$ can take to the
other input terminal representing concept ``$x_i$ is small''. By
this trick, when the value of variable $x_i$ is small, applied value
to the input terminal representing concept ``$x_i$ is small'' will
be high showing that our confidence degree about the validity of the
concept ``$x_i$ is small'' is high. Similarly, in the application of
edge detection, intensity value of pixel will be directly applied to
input terminal representing concept ``pixel is bright'' and its
negative (255 minus the intensity value of the pixel) will be
applied to input terminal representing concept ``pixel is dark''.

Now that we have proposed a method to construct fuzzy concepts at
the input stage of our system, it is time to define the shape of the
membership functions of these fuzzy sets (concepts). For this
purpose, a simple preprogrammed memristor crossbar structure is
considered in the fuzzification unit of the circuit of Fig.
\ref{fuzzyXOR} which is inspired from the first layer of the network
of Fig. \ref{fig1b}. In this memristor crossbar structure, the first
two vertical wires somehow acts as a universe of discourse of
variable $x_1$ while the next two vertical wires represent the
universe of discourse of the other variable, {\it i.e.} $x_2$. In
this case, preprogrammed weights at crosspoints of the crossbar will
have the role of the membership functions of the fuzzy sets defined
on the universe of discourses of input variables. In fact, it is the
configuration and value of these weights that specify the shape of
these membership functions. For example, In the fuzzification unit
of the circuit of Fig. \ref{fuzzyXOR}, weights on the first and
second columns of the crossbar define the shape of the membership
functions of fuzzy sets ``big'' and ``small'' respectively on the
universe of discourse of variable $x_1$. The shape of these
membership functions that we have considered in this sample circuit
is depicted in the left side of the crossbar while their numerical
specifications are presented in the inset of the figure. Note that
by the reprogramming of memristors at crosspoints of the crossbar,
shapes of these membership function and their support can be simply
changed. We have a similar case for input variable $x_2$. It is
evident that any number of fuzzy sets with any arbitrary membership
functions can be implemented in a similar way.

Now, lets see how the fuzzification unit of our proposed system
works. To clarify the working procedure of this unit, consider the
following simple case. Assume that the current values of input
variables $x_1$ and $x_2$ are $x_1^{obs}$ and $x_2^{obs}$
respectively. In this case, similar to what we had in Section
\ref{sec22}, output column vector of the fuzzification unit, {\it
i.e.} $\mathbf{v}_{fuzzification}$, can be written as:
\begin{eqnarray}\label{eq4}
\mathbf{v}_{fuzzification}=x_1^{obs}\mathbf{s}_1+\left(x_1^{max}-x_1^{obs}\right)\mathbf{s}_2+x_2^{obs}\mathbf{s}_3+\left(x_2^{max}-x_2^{obs}\right)\mathbf{s}_4
\end{eqnarray}
where $\mathbf{s}_i$ for $i=1,2,3,4$ is the column vector
representing predetermined weights on the $i$th column of the
crossbar located between the input and the first hidden layer and
$x_i^{max}$ for $i=1,2$ is the maximum value that input variable
$x_i$ can take. However, since weights are programmed on the first
two columns of the crossbar in a way that they do not have overlap
with weights on the next two columns, it can be said that upper rows
of this crossbar creates the weighted sum of the membership
functions defined on the universe of discourse of input variable
$x_1$ while lower rows of the crossbar generates the weighted sum of
those membership functions which are defined on the universe of
discourse of input variable $x_2$. Therefore, on the upper rows of
the first hidden layer (output of the fuzzification unit) we will
have a fuzzy number with the membership function of
$x_1^{obs}\mathbf{s}_1+\left(x_1^{max}-x_1^{obs}\right)\mathbf{s}_2$
corresponding to the applied crisp input value $x_1^{obs}$ and on
the lower rows we will have a fuzzy number with the membership
function of
$x_2^{obs}\mathbf{s}_3+\left(x_2^{max}-x_2^{obs}\right)\mathbf{s}_4$
corresponding to the applied crisp input value $x_2^{obs}$.

To summarize, the role of the fuzzification part of the circuit is
to convert the crisp input numbers to their corresponding fuzzy
numbers where the shape of these fuzzy numbers are specified through
the weights which are programmed on columns of the crossbar.
Finally, note that output of this part of the circuit are from a
kind of membership degrees and therefore they are always
non-negative.

\subsection{the fuzzy minterm creating unit}

That part of the network of Fig. \ref{fuzzyXOR} which is located
between the first and the second hidden layer and specified by a
grayscale dashed rectangle is called the fuzzy minterm creating
unit. The main role of this part of the circuit is to compare the
created fuzzy numbers by the fuzzification unit with some patterns
which are programmed on the columns of the crossbar. In fact, this
section of the proposed system performs a dot product between input
fuzzy numbers (output of the first hidden layer) and the weight
vectors which are formed on columns of the middle crossbar.
Therefore, it somehow measures the available similarities between
input fuzzy numbers and pre-programmed weights on columns of the
crossbar.

However, the fuzzy minterm creating unit actually does something
more than a simple dot product between vectors. To make it more
clear, consider the first (the left-most) column of the crossbar in
this unit. The output of its corresponding neuron (connected to this
column) will be maximum only when the both of input variables, {\it
i.e.} $x_1$ and $x_2$, have their maximum values (Note that in this
structure, we always have this condition that both of the input
variables and weight vectors are non-negative). Therefore, it can be
said that this column of the crossbar implements the antecedent part
of the following fuzzy rule:

IF $\quad$ $x_1$ is \textit{big} $\quad$ AND $\quad$ $x_2$ is
\textit{big} $\quad$ THEN $\quad$ $x_1 \oplus x_2$ is \textit{small}

and output of its corresponding neuron for any observed inputs will
show the result of the evaluation of the antecedent part of this
rule for this given data. Since inputs of this unit of the system
are fuzzy numbers and this column of the crossbar somehow implements
the AND function between input variables themselves and not their
complements, we can assume that it creates the first fuzzy minterm,
{\it i.e.} minterm number 0. Similar to Boolean minterms, this
minterm will take its maximum output value if and only if both of
its input variables have their maximum values (or equivalently when
both of them are ``big''). In a similar way, the second column of
the crossbar clearly implements the antecedent part of the following
fuzzy rule:

IF $\quad$ $x_1$ is \textit{big} $\quad$ AND $\quad$ $x_2$ is
\textit{small} $\quad$ THEN $\quad$ $x_1 \oplus x_2$ is \textit{big}

which its output can be considered as the second fuzzy minterm
between input variables (minterm number 1). This is because of the
fact that this column implements the AND function between the
variable $x_1$ (or concept ``$x_1$ is big'') and the complement of
the second variable $x_2$ (or concept ``$x_2$ is small'').
Therefore, for any different configuration of two parts of the
antecedent parts of the rules, one distinct fuzzy minterm will be
constructed. These fuzzy minterms have this property that at anytime
and based on input data, all of them will be active with different
strengths but only one of them will have higher output value
compared with other minterms. As a result, by this way we can
recognize which concepts have been happened simultaneously.

Now, let's assume that specific inputs are applied to the system and
we want to evaluate the antecedent part of each of these rules for
these inputs. In the other words, for these inputs, we want to
determine the strength of the activation of each of these fuzzy
minterms. Actually, this evaluation task is the second main duty of
the fuzzy minterm creating unit which is done in this part of our
system through the dot product of the membership functions of the
created fuzzy numbers (outputs of the first hidden layer) and weight
vectors programmed on the columns of the crossbar.

Finally, note that based on the provided explanations about the
thresholding task in artificial neural networks, the activation
function of the output neurons of the fuzzy minterm creating unit is
considered to be $x^n$ to magnify the difference between outputs of
neurons of this layer. However, the role of this kind of activation
function can be interpreted in another way; it somehow emulates the
role of the AND operation between two parts of the antecedent of
fuzzy rules. Note that we usually connect parts of the antecedent by
a conjunction (`AND') to have a simple way to know when these two
parts happen simultaneously. If these two parts happen at the same
time, the evaluation result of the corresponding rule will be higher
than any other rules. By the use of the $x^n$ ($n>1$) we will have
the same condition. In this case, evaluation result of those rules
(fuzzy minterms) which have only one active part in their antecedent
will be much less than the evaluation result of the rule which both
parts of its antecedent are active at the same time. Actually, here
we tied to reveal the similarities between fuzzy rule bases and
truth tables in digital logic.

\subsection{The aggregation unit}

The last part of the circuit of Fig. \ref{fuzzyXOR} is the
aggregation unit located between the second hidden layer and the
output layer. The main role of this unit is to aggregate consequence
parts of the rules based on their evaluation results. In our
proposed structure, this process is done by summing outputs of those
rules (neurons) which have the same consequence part. Therefore, we
will have one output per each different consequence part (concept)
in fuzzy rule base. Here, some differences are visible between our
proposed inference system and other common inference methods. First
of all, we have used a summation operator as a triangular conorm to
aggregate fuzzy rules. Second, since our primary goal was to
construct a system with fuzzy input and fuzzy output terminals, no
defuzzification unit is intended at the output stage of the system.
Consequently, as mentioned before, at the output stage of our
system, instead of one simple crisp output, we will have one output
per each distinct consequence part (output concept). Herein, it
should be noted that similar to other t-conorm operators, the result
of this summing operation is of a kind of membership degrees which
determines confidence degrees about the validity of those concepts
which are assigned to output neurons. For example, in the network
depicted in Fig. \ref{fuzzyXOR}, the first output neuron represents
concept ``$x_1\oplus x_2$ is big''. In this case, by the increase of
the output of this neuron, our belief about the occurrence of the
concept ``$x_1\oplus x_2$ is big'' which is assigned to this neuron
increases as well. Since the concept ``$x_1\oplus x_2$ is big'' is
the consequence part of these following rules:

IF $x_1$ is big AND $x_2$ is small THEN $x_1\oplus x_2$ is big

IF $x_1$ is small AND $x_2$ is big THEN $x_1\oplus x_2$ is big

which are constructed on the second and the third columns of the
crossbar of the fuzzy minterm creating unit, outputs of their
corresponding neurons are summed together in the aggregation unit to
create a single output neuron for representing concept ``$x_1\oplus
x_2$ is big''. In this case, when the value of the XOR function
between input variables $x_1$ and $x_2$ is high ($x_1$ and $x_2$
differ significantly), output of this neuron will be more than any
other neurons at the output layer. As a result, since the output of
this neuron in the aggregation unit has a direct relationship with
the result of the XOR function between input variables, it can be
simply considered as a final output of the system. Therefore, unlike
to traditional fuzzy systems which usually consider the own concepts
such as ``y=1'' or ``y=10'' as their final outputs, output of our
system is the confidence degrees of these concepts.

Finally, it should be emphasized that for each output concept in the
circuit of Fig. \ref{fuzzyXOR}, we have considered only one single
row (output). This is because of the fact that by this way, the
output of the system will become a single number. However, it is
clear that by increasing the number of rows for each concept and
programming them properly, it is possible to get fuzzy numbers for
each output concept as well.

In the next section, we will describe how this circuit can be used
as an image processing system to extract edges from any given
grayscale image.

\section{application of the constructed fuzzy XOR function for doing edge detection in grayscale images}
\label{edge} In this section, we want to show how our proposed fuzzy
XOR gate can be applied to grayscale images to extract edges from
them. First of all, let's look at the process of edge detection in
binary images briefly. In this kind of images, edges are located
between pixels with different intensities. In the other words,
wherever we have one black (with low intensity) and one white (with
high intensity) pixel near each other, we will have edge between
them. Otherwise, when neighbor pixels have the same pixel values, no
edge will exist. Consequently, edges can be detected between
neighbor pixels in binary images by the application of the logical
XOR function to them: when neighbor pixels have different
intensities, the output of the logical XOR function will be high
(logic 1) which is the indicator of the existence of an edge between
these pixels.

\begin{figure}[!t]
\centering
\includegraphics[width=7in,height=5.5in]{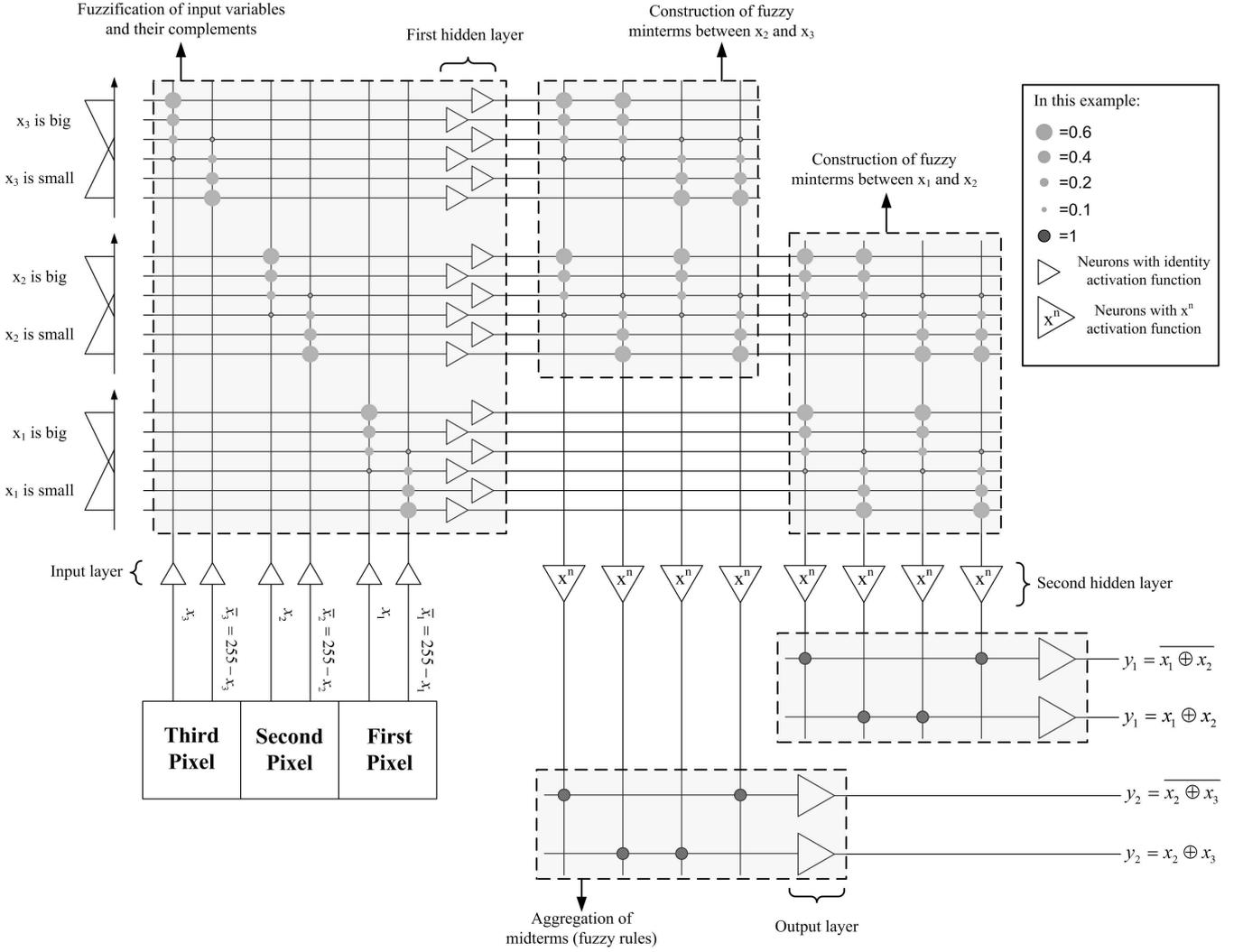}
\caption{The result of merging two fuzzy XOR systems of Fig.
ref{fuzzyXOR} to extract edges from three consecutive pixels.}
\label{edgestruct} 
\end{figure}

Now, consider the case in which we have grayscale images instead of
binary ones. In order to detect edges in this kind of images, we
need a simple function like the binary XOR gate but by this
difference that it should be able to work with continuous values
instead of binary numbers. On the other hand, this function should
also have the following simple property: when both of input pixels
have similar intensities, output of this function should be near
zero but by the increase of the difference between the intensities
of input pixels, the output of this function should increase as
well. Therefore, output of this function or system should be
directly proportional to the difference between the intensity values
of input pixels. If we can design and build a function with these
properties, then we can apply it to the consecutive pixels of the
input grayscale image and get an image as the output result where in
that image edges are specified proportional to their strengths:
stronger edges have higher pixel values than weaker ones. One of
such systems which has this property is our proposed circuit
depicted in Fig. \ref{fuzzyXOR}. It generates low(high) output when
its input variables have similar(different) values. That is why we
have called our proposed system in Section \ref{fuzzyxorgate}
``fuzzy XOR gate''. It is clear that in this structure, unlike
binary XOR function, input and output variables can be continuous.
However, our proposed fuzzy XOR structure has this advantage that it
can extract all horizontal or vertical edges simultaneously. This is
because of the fact that it is possible to use several of these
fuzzy XOR gates at the same time. Without the loss of generality,
Fig. \ref{edgestruct} shows the process of extracting edges from
three neighbor pixels. In this circuit, two fuzzy XOR gates are
merged to each other to optimize the overall system. In the output
layer, we have one output per each pair of consecutive pixels
showing the result of the fuzzy XOR function for these pixels. By
adding more fuzzy XOR gates to this system in a same manner, we can
construct a structure which can extract all vertical edges in one
row of the image simultaneously. By using similar circuits for other
rows of the image, all vertical edges in the entire image can be
extracted. At the same time, by rotating the image by 90 degrees and
repeating the same procedure, horizontal edges can be extracted as
well. Note that since the circuit of Fig. \ref{fuzzyXOR} or Fig.
\ref{edgestruct} is in analog form, it can detect edges (do fuzzy
inference) in real-time.

\section{Simulation results}
\label{simulation} In this section, we will illustrate the
efficiency and applicability of our proposed method (for the
hardware implementation of fuzzy inference systems like the fuzzy
XOR function) by performing several experiments. In all of the
following simulations, the structure of Fig. \ref{edgestruct} is
used with the same pre-programmed membership functions on the
crossbars with those numerical specifications given in the inset of
the figure. In addition, since outputs of the system may not be
bounded between 0 and 255 (as required by the grayscale images),
output images are mapped to this range before presenting in this
paper. However, it should be noted that it would be easy to modify
the specifications of the system of Fig. \ref{edgestruct} ({\it
e.g.} opamps gains, shapes of the membership functions and etc.) in
order to force it to create outputs between 0 and 255. Moreover, to
remove probable noises in input images, all inputs are smoothed with
the Gaussian smoothing filter before being applied to the system.

\begin{figure}[!t]
\centering \subfigure[]{
\label{firstsim:a} 
\includegraphics[width=3.1in,height=1.9in]{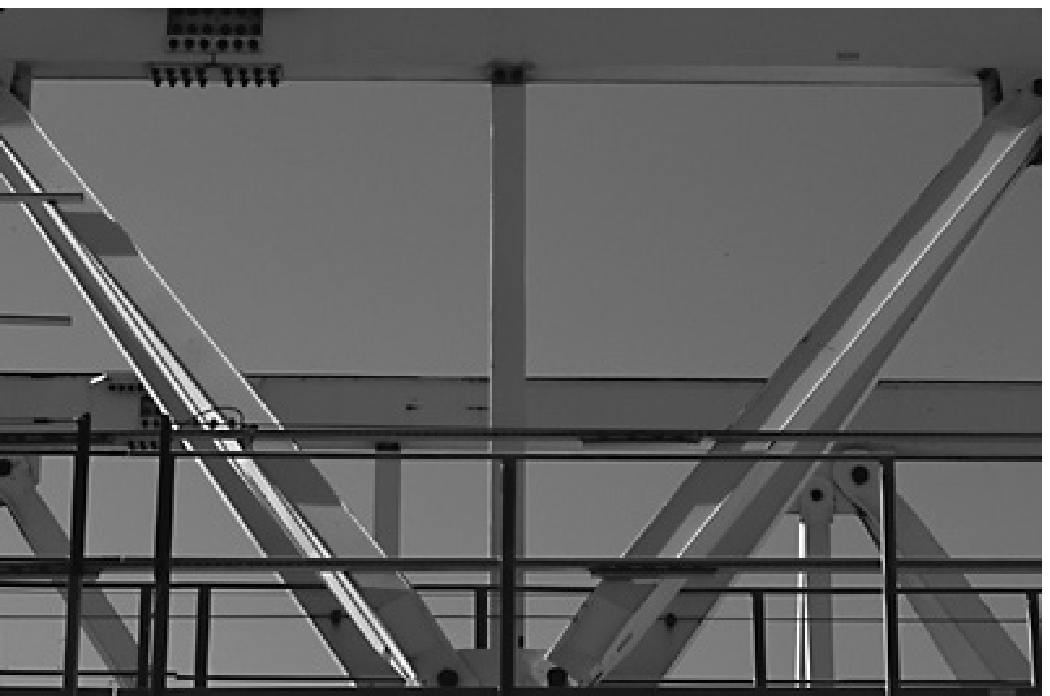}}
\vspace{0.05in} \subfigure[]{
\label{firstsim:b} 
\includegraphics[width=3.1in,height=1.9in]{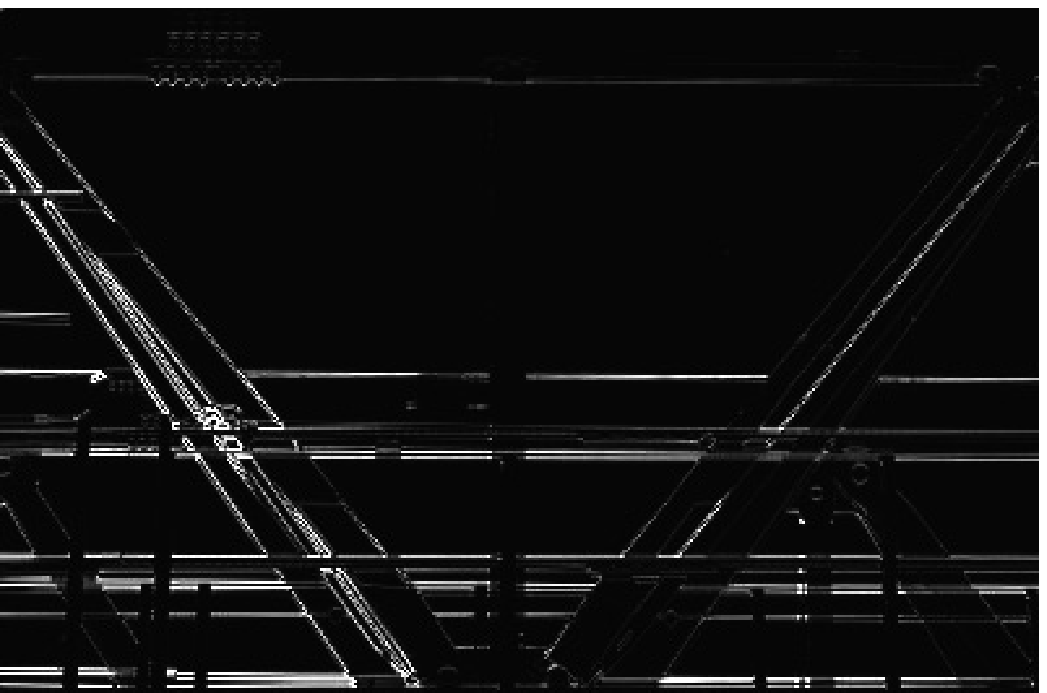}}
\subfigure[]{
\label{firstsim:c} 
\includegraphics[width=3.1in,height=1.9in]{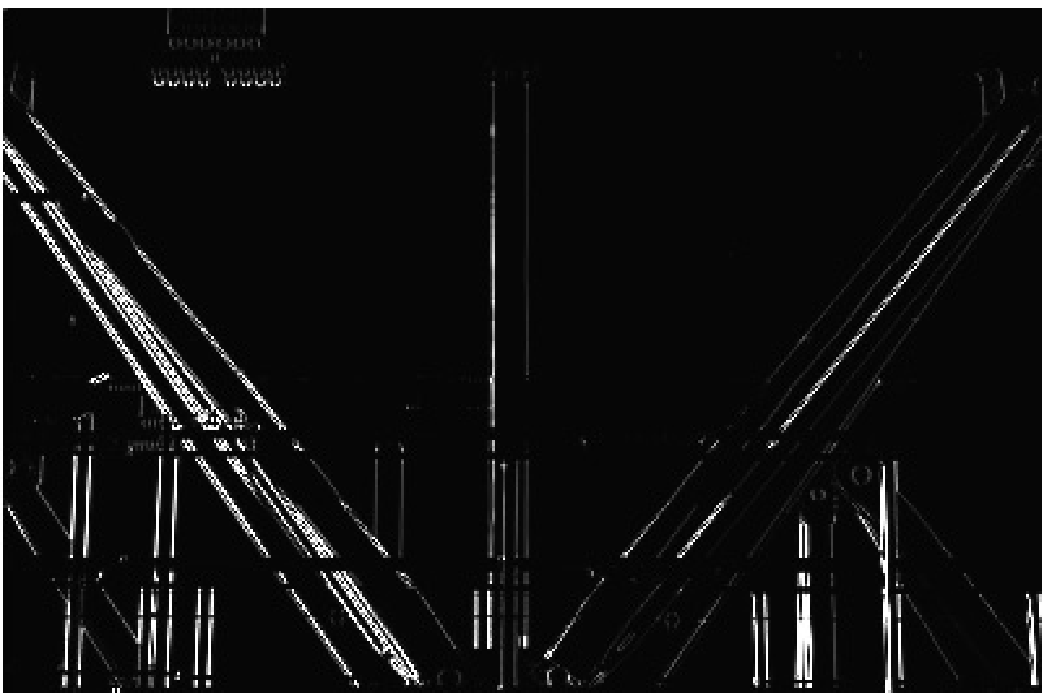}}
\subfigure[]{
\label{firstsim:d} 
\includegraphics[width=3.1in,height=1.9in]{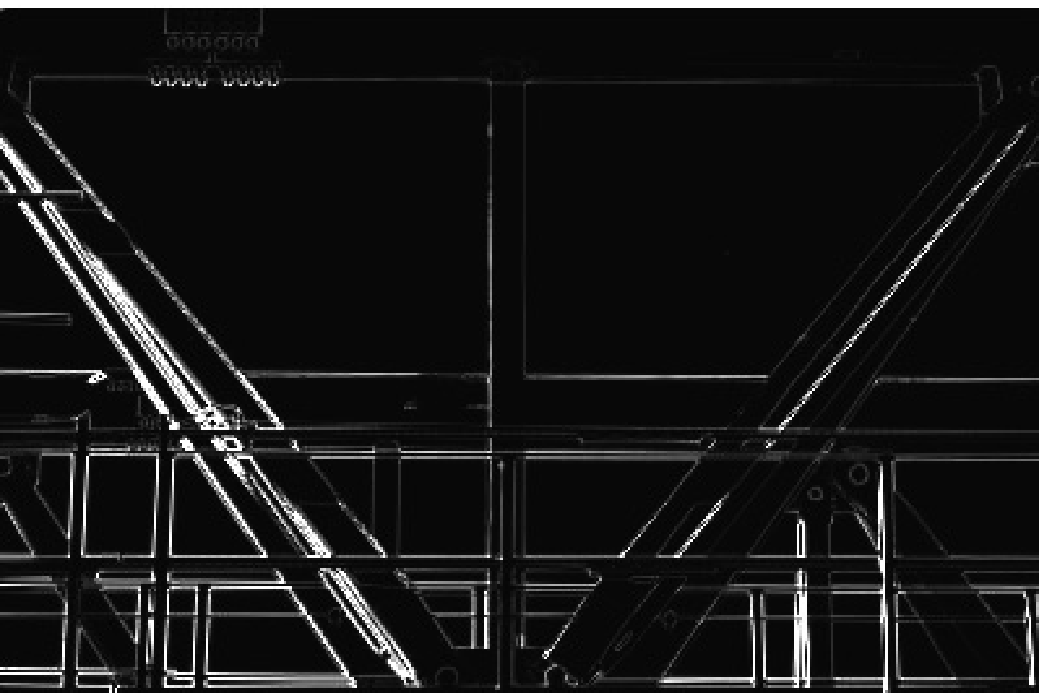}}
\subfigure[]{
\label{firstsim:e} 
\includegraphics[width=3.1in,height=1.9in]{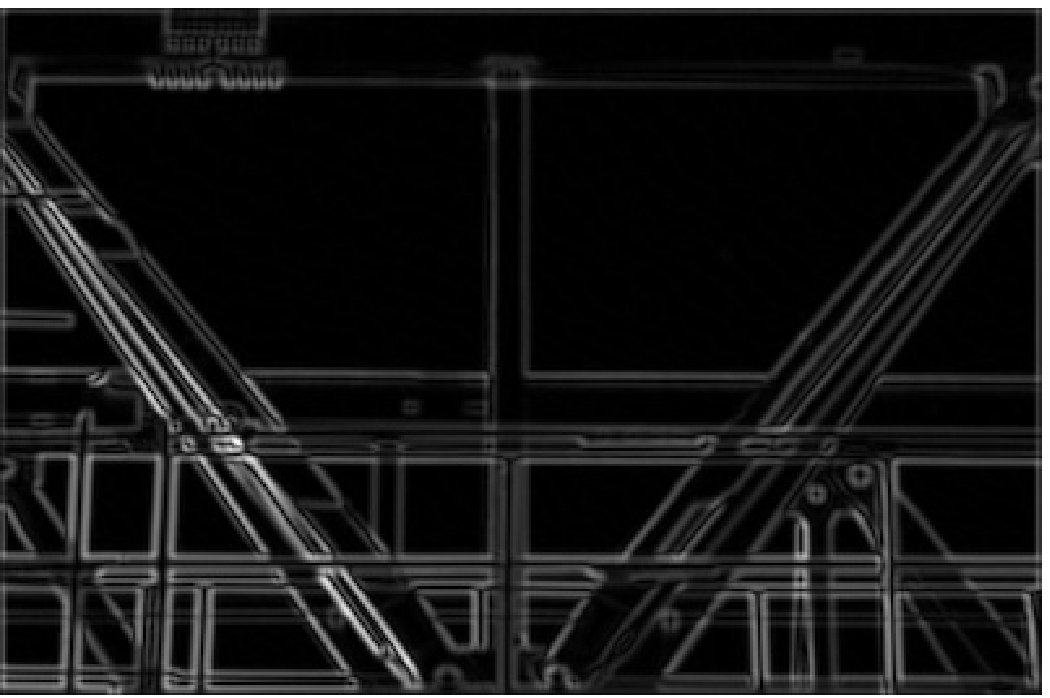}}
\subfigure[]{
\label{firstsim:f} 
\includegraphics[width=3.1in,height=1.9in]{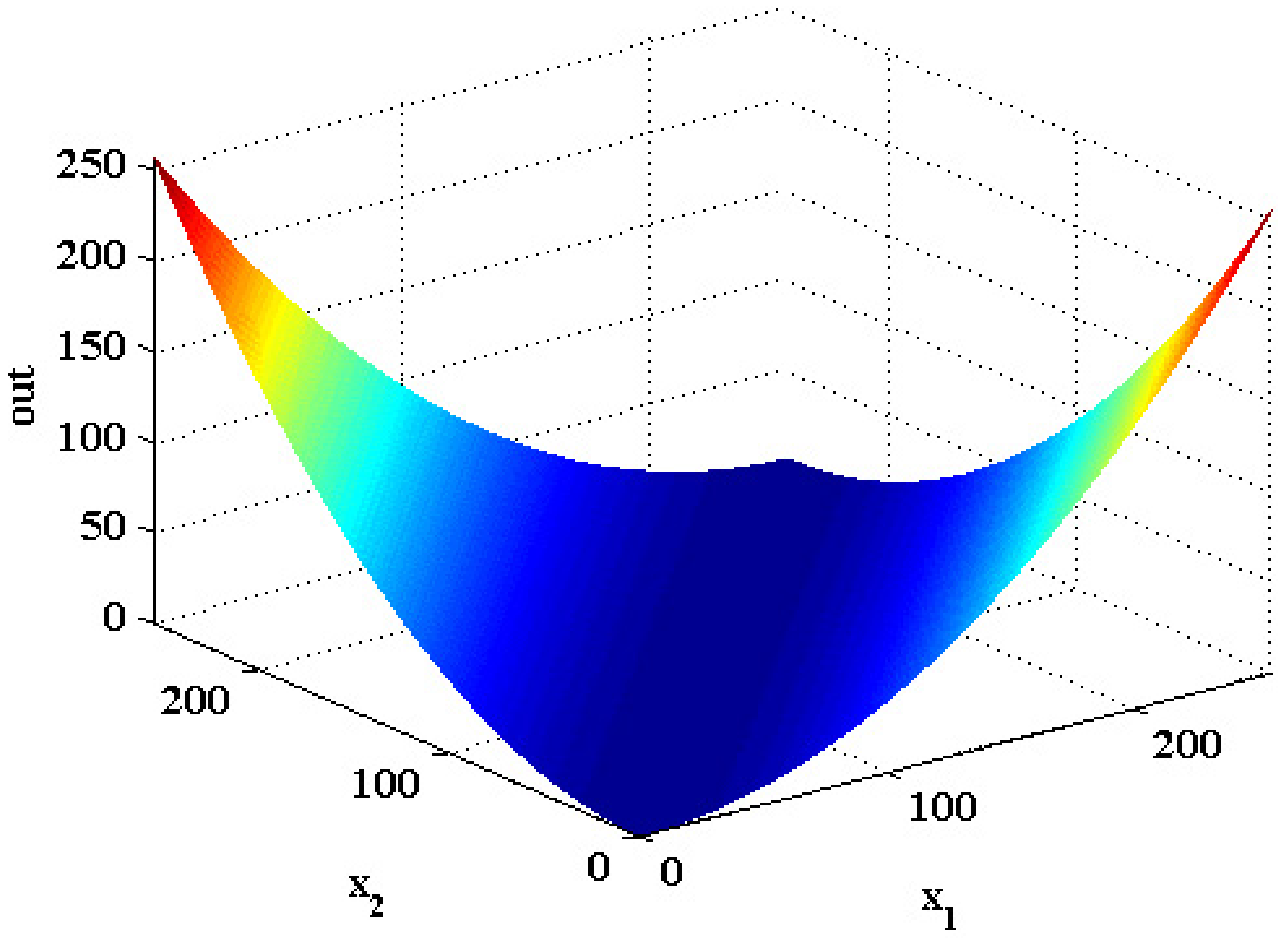}}
\caption{simulation results of the first conducted experiment. (a)
Input image. (b) Extracted horizontal edges by using our method. (c)
Extracted vertical edges by our proposed system. (d) Final output of
our fuzzy XOR function. (e) Extracted edges by applying the first
two steps of the canny edge detection algorithm. (f) Fuzzy inference
surface of the system of Fig. \ref{fuzzyXOR}.}
\label{firstsim} 
\end{figure}

The results of the first conducted simulation is presented in Fig.
\ref{firstsim}. In this simulation, the image shown in Fig.
\ref{firstsim:a} is applied to the system of Fig. \ref{edgestruct}
as an  input and extracted horizontal and vertical edges are
presented in Figs. \ref{firstsim:b} and \ref{firstsim:c}
respectively. By merging these two images, one single image can be
obtained as a final result of our proposed edge detection algorithm.
Figure \ref{firstsim:d} shows this image for the given input image
of Fig. \ref{firstsim:a} which is obtained simply by adding two
images of Figs. \ref{firstsim:b} and \ref{firstsim:c}. This figure
shows that our proposed circuit can effectively extract edges from
grayscale images. It also indicates that output images of this
structure all have this property that their intensities at any point
are directly proportional to the strength of the existing edges at
that point in the original input images. Note that since outputs of
the system of Fig. \ref{edgestruct} are always non-negative (because
they are of a kind of confidence degrees), horizontal and vertical
edges can be directly summed without any concerns and therefore the
application of the Manhattan distance measure is not necessary
anymore. In order to have better view about the performance of the
proposed method, the result of the first two steps of the canny edge
detection algorithm \cite{canny}, {\it i.e. smoothing and finding
gradients}, applied to the image of Fig. \ref{firstsim:a} is shown
in Fig. \ref{firstsim:e}. By comparing Fig. \ref{firstsim:d} with
Fig. \ref{firstsim:e}, it can be inferred that although the input
image is smoothed, our structure has produced sharper edges than its
counterpart in the canny edge detection algorithm. In addition,
especially in those areas of Fig. \ref{firstsim:a} where the image
is uniform, unwanted detected edges are abundant and visible in Fig.
\ref{firstsim:e} which is not the case in Fig. \ref{firstsim:e}.
Note that our proposed method has also this advantage versus most of
other edge detecting algorithms that as demonstrated in this paper,
it can be implemented efficiently in analog form and therefore it
can easily operate in real-time. Finally, Fig. \ref{firstsim:f}
shows the fuzzy inference surface of the system of Fig.
\ref{fuzzyXOR} which is obtained by applying different values of
input variables (between 0 and 255) to the system and plotting its
outputs versus these input values. In this figure, the overall
behavior and shape of the fuzzy XOR function is clearly observable:
in those areas where input variables have similar values (pixels
have similar intensities), output of the system is near to zero.
However, by the increase of the difference between values of input
variables, output of the system begins to approach its upper bound,
{\it i.e.} 255. Although this figure is obtained by using the $x^2$
activation function for neurons of the second hidden layer of the
circuit of Fig. \ref{edgestruct}, our experiments showed that using
other similar activation functions like $x^4$ or $x^7$ has little
impact on the shape of this fuzzy inference surface. It is also
interesting to know that by the use of Mamdani's fuzzy inference
method \cite{mamdani} or Takagi-Sugeno-Kang method of fuzzy
inference \cite{takagi1,takagi2}, it is not possible to create such
a surface from the aforementioned fuzzy rule bases describing the
fuzzy XOR function.

\begin{figure}[!t]
\centering \subfigure[]{
\label{secondsim:a} 
\includegraphics[width=2in,height=2in]{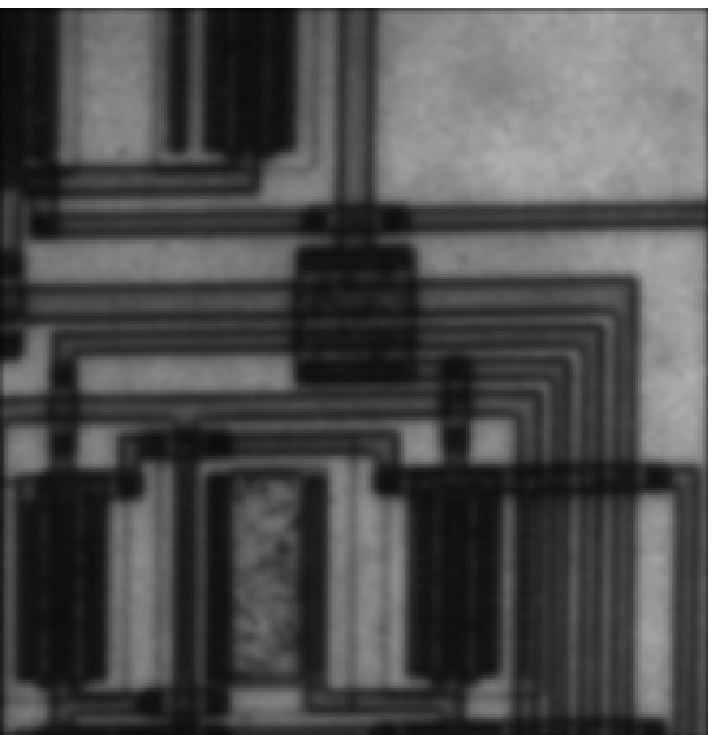}}
\vspace{0.05in} \subfigure[]{
\label{secondsim:b} 
\includegraphics[width=2in,height=2in]{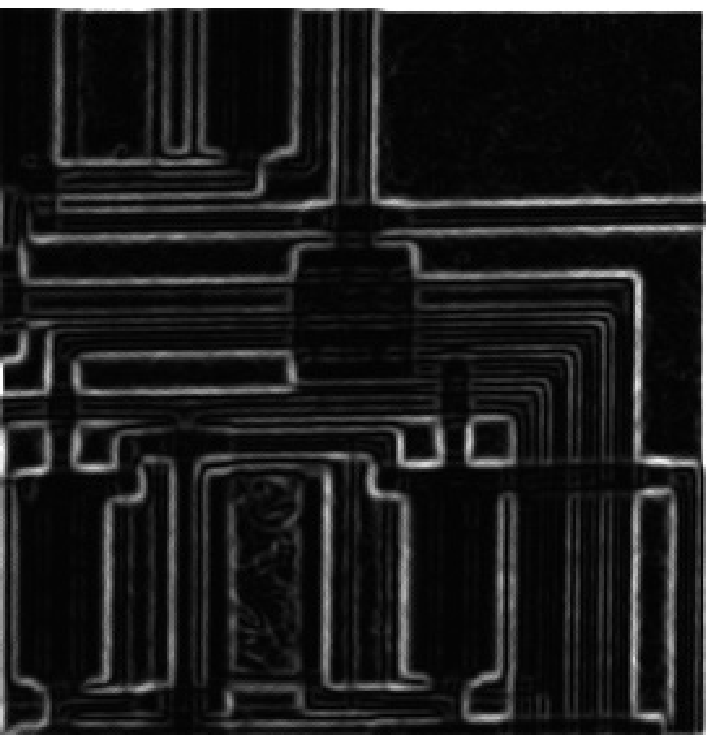}}
\subfigure[]{
\label{secondsim:c} 
\includegraphics[width=2in,height=2in]{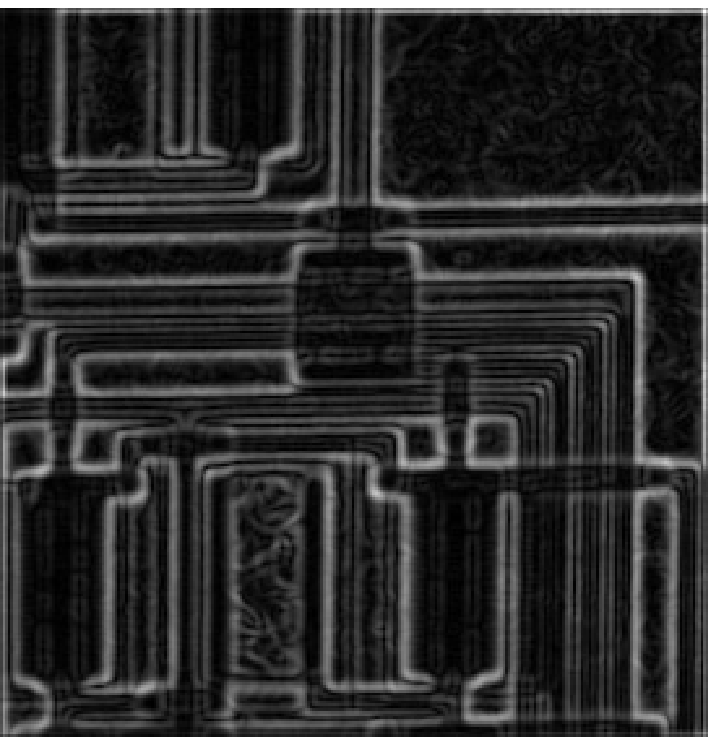}}
\subfigure[]{
\label{secondsim:d} 
\includegraphics[width=2in,height=2in]{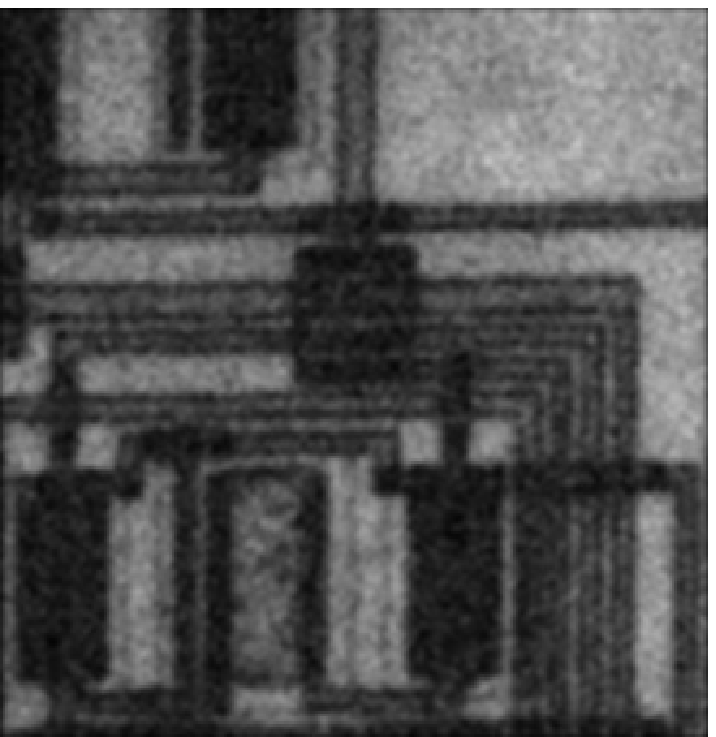}}
\subfigure[]{
\label{secondsim:e} 
\includegraphics[width=2in,height=2in]{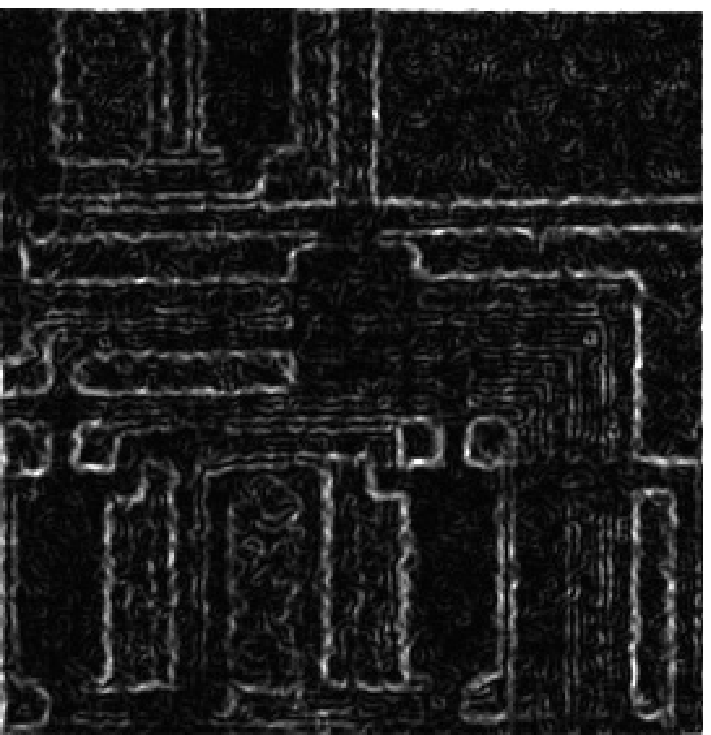}}
\subfigure[]{
\label{secondsim:f} 
\includegraphics[width=2in,height=2in]{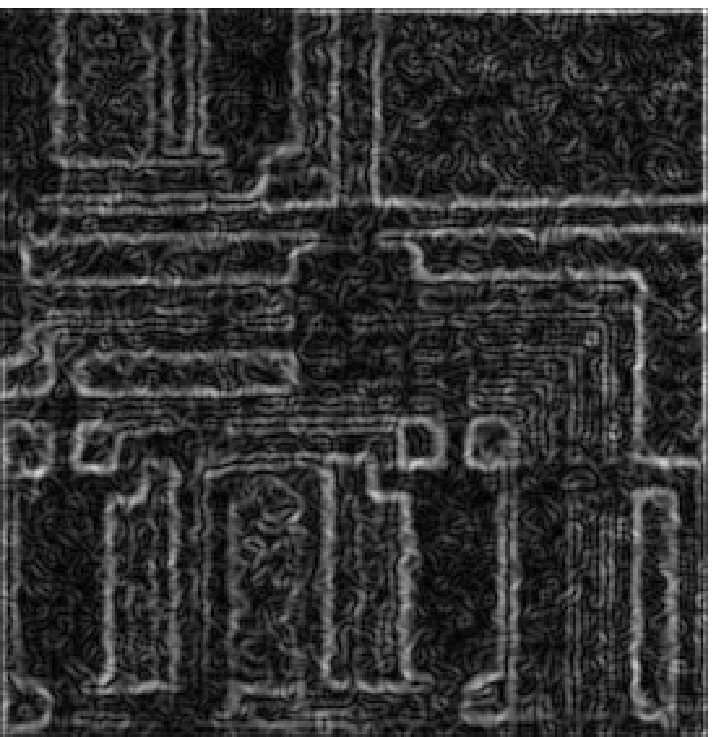}}
\subfigure[]{
\label{thirdsim:a} 
\includegraphics[width=2in,height=2in]{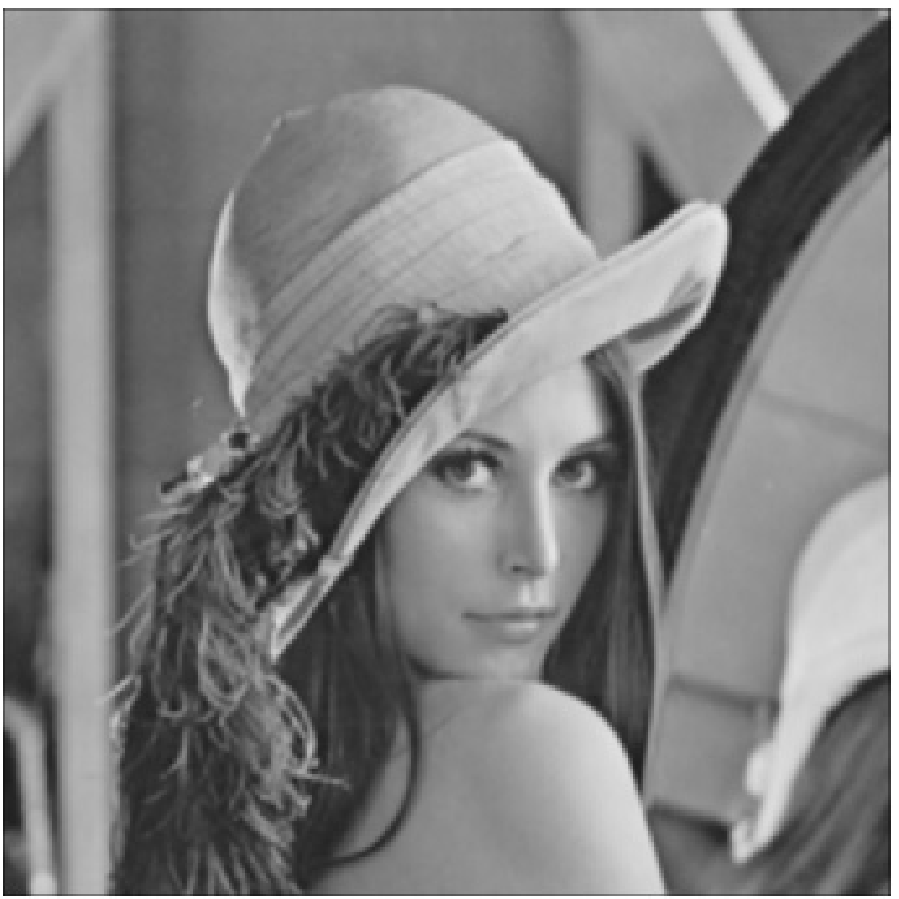}}
\vspace{0.05in} \subfigure[]{
\label{thirdsim:b} 
\includegraphics[width=2in,height=2in]{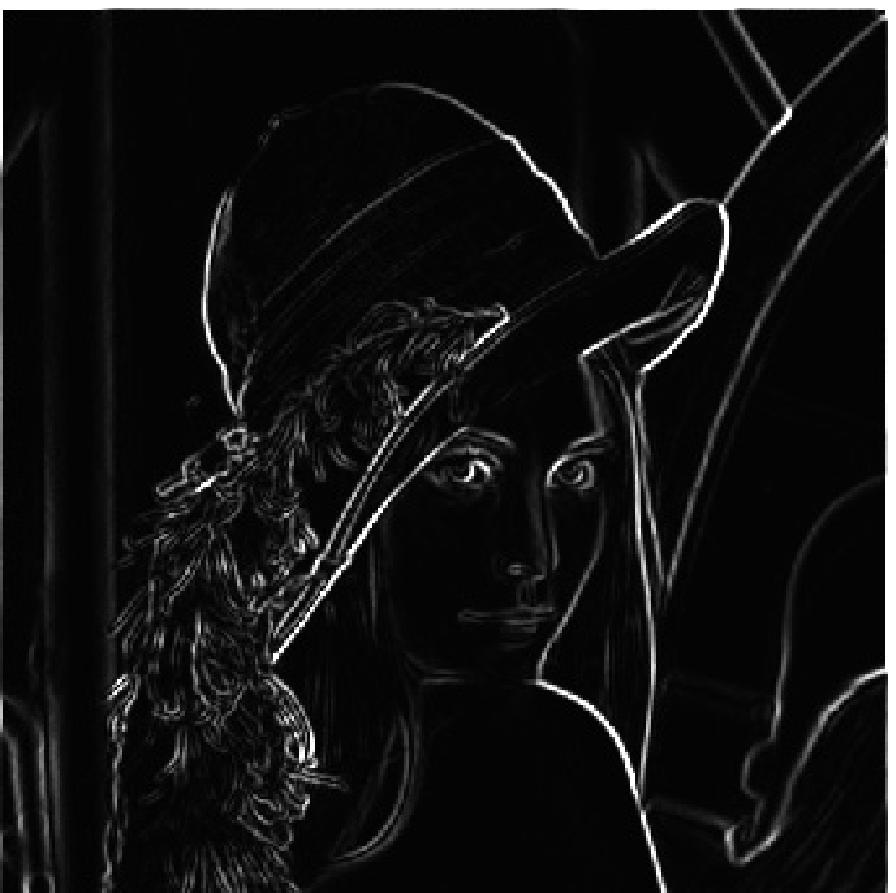}}
\subfigure[]{
\label{thirdsim:c} 
\includegraphics[width=2in,height=2in]{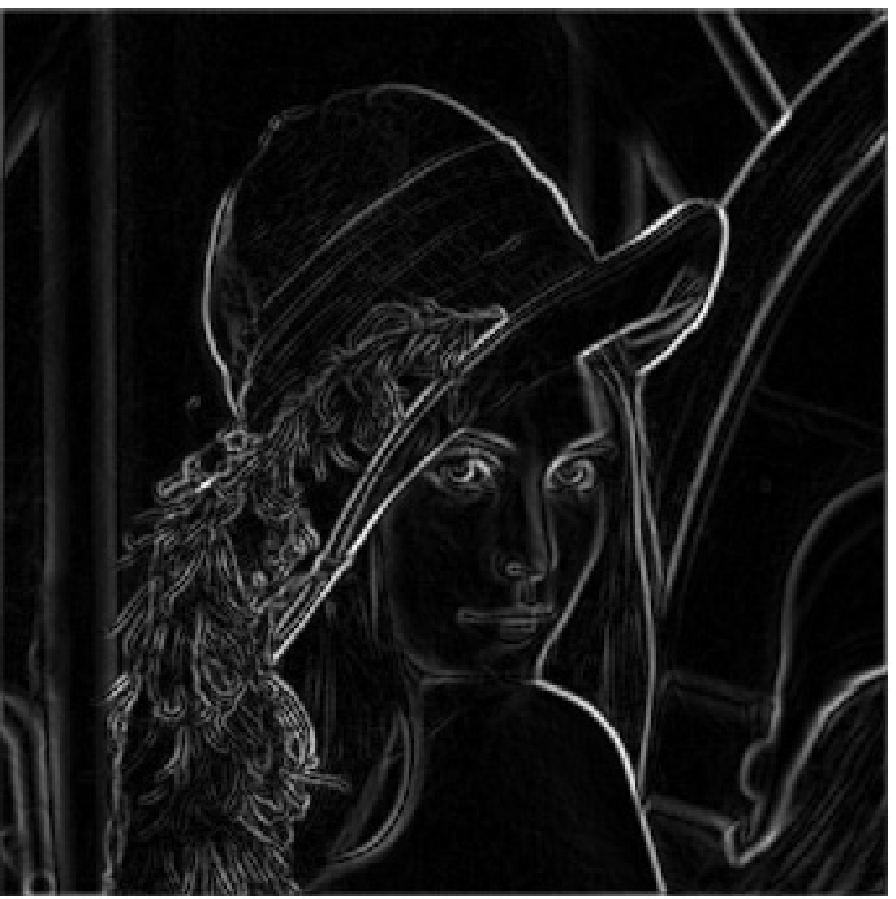}}
\subfigure[]{
\label{thirdsim:d} 
\includegraphics[width=2in,height=2in]{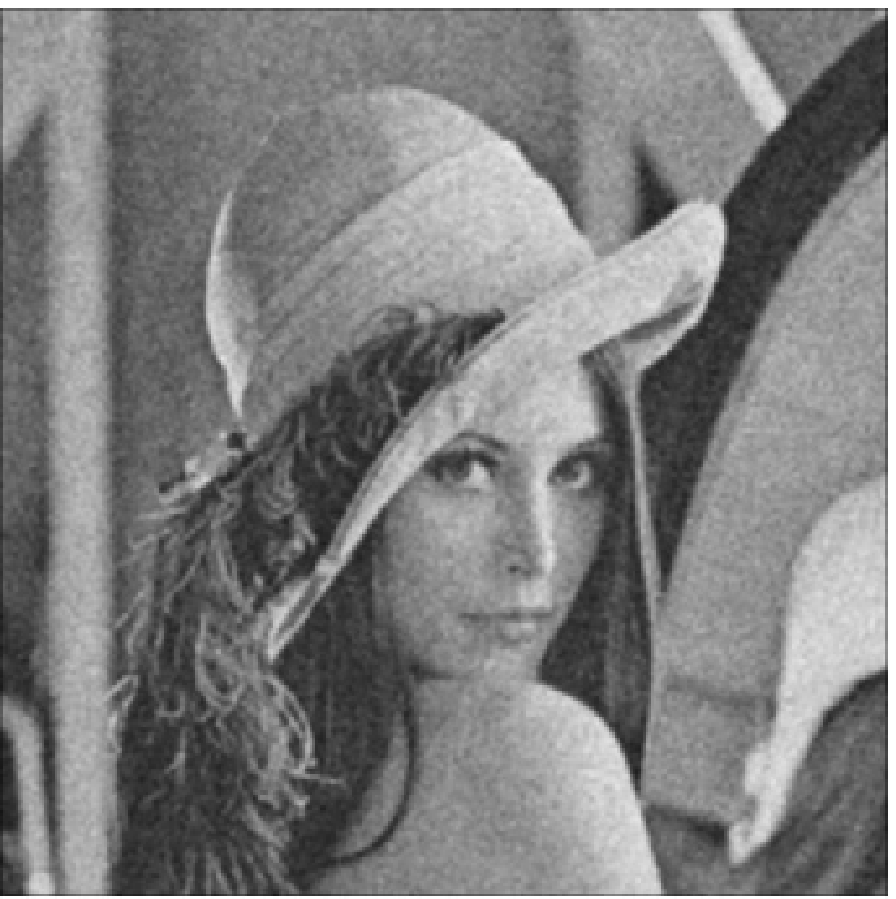}}
\subfigure[]{
\label{thirdsim:e} 
\includegraphics[width=2in,height=2in]{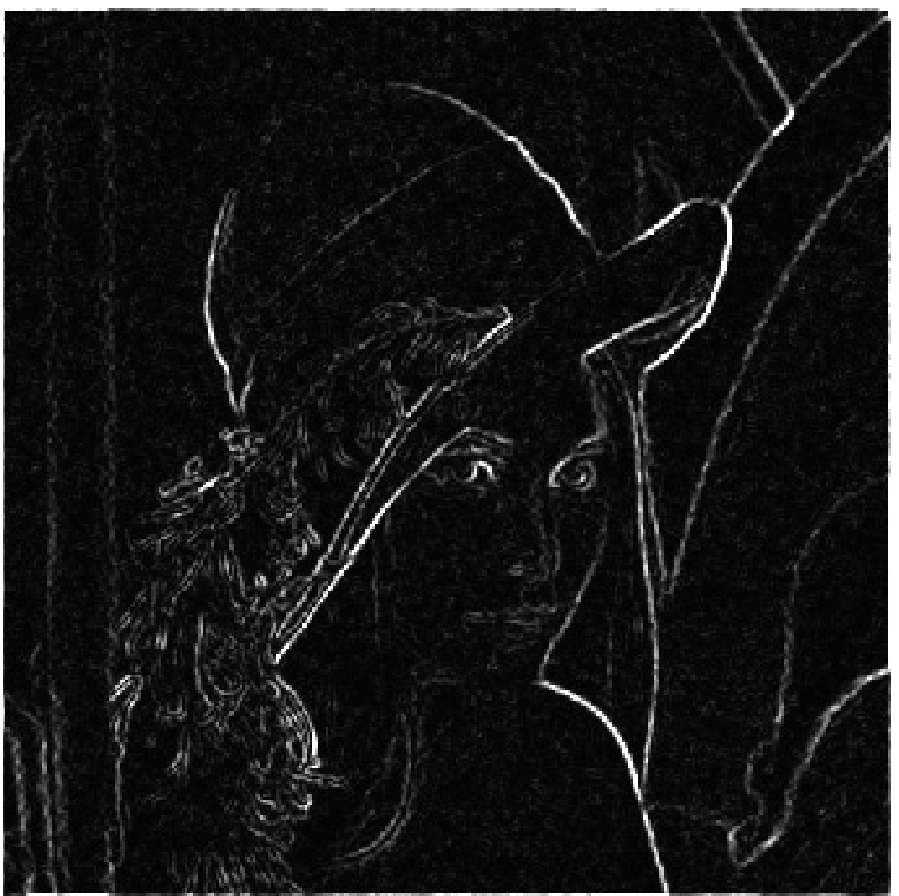}}
\subfigure[]{
\label{thirdsim:f} 
\includegraphics[width=2in,height=2in]{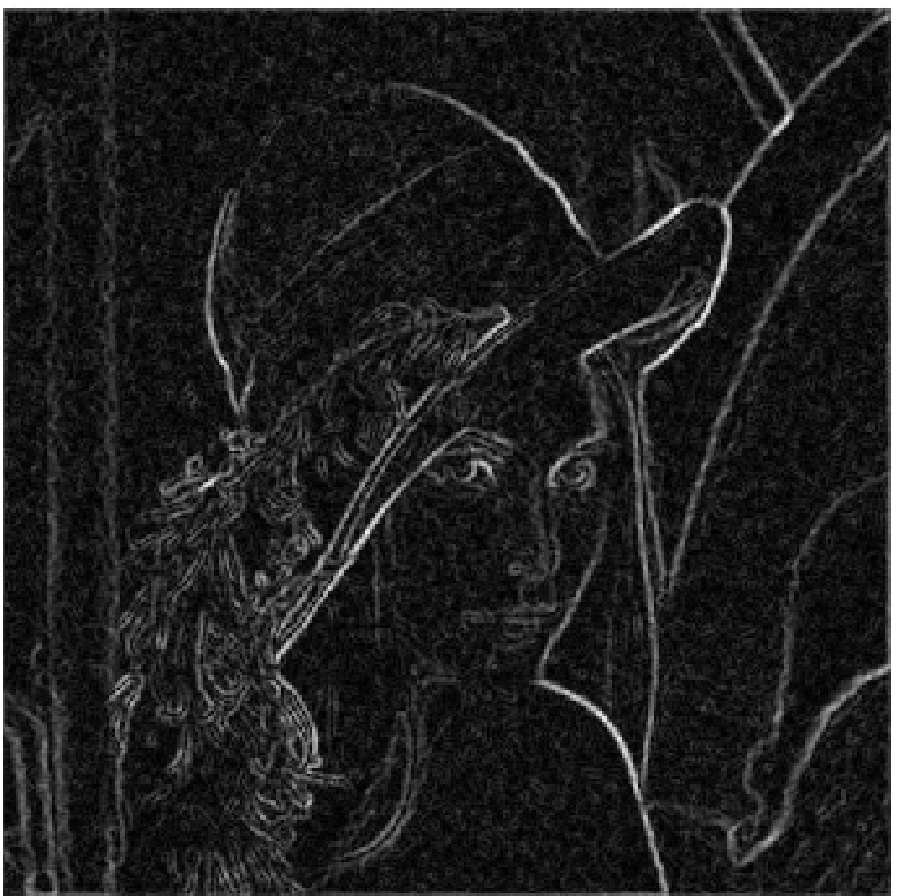}}
\caption{simulation results of the second conducted experiment. (a
and g) Input images. (b and h) Extracted edges by using our proposed
structure. (c and i) Extracted edges by applying the first two steps
of the canny edge detection algorithm. (d and j) Input images
degraded by Gaussian noise. (e and k) Extracted edges from noisy
input by using our proposed structure. (f and l) Extracted edges
from noisy input by applying the first two steps of the canny edge
detection algorithm.}
\label{secondsim} 
\end{figure}

In the next simulation, we used two different images as an input and
applied our proposed structure to extract edges from them. These
input images are shown in Fig. \ref{secondsim:a} and Fig.
\ref{thirdsim:a}. Extracted edges from these figures by using our
method and structure are presented in Fig. \ref{secondsim:b} and
Fig. \ref{thirdsim:b} and the result of applying the first two steps
of the canny edge detection algorithm to these input images are
demonstrated in Fig. \ref{secondsim:c} and Fig. \ref{thirdsim:c}. To
illustrate the stability and performance of our memristive fuzzy XOR
system in noisy environment, detected edges from noisy images of
Fig. \ref{secondsim:d} and Fig. \ref{thirdsim:d} which are obtained
by adding Gaussian white noise of mean 0 and variance 0.03 to input
images are shown in Fig. \ref{secondsim:e} and Fig.
\ref{thirdsim:e}. To have better comparison, the result of applying
the smoothing and finding gradients steps of the canny edge
detection algorithm to these noisy images are presented in Fig.
\ref{secondsim:f} and Fig. \ref{thirdsim:f}. From the result of this
simulation, it can be inferred that although our proposed edge
detection method only uses information of two neighbor pixels, it
performs acceptably against noisy images.

\section{conclusion}
\label{conclusion} In this paper we proposed a new hardware based on
memristor crossbar structure to implement a fuzzy edge detector
algorithm. For this purpose, at first we expressed fuzzy XOR
function in the form of fuzzy rule base and then implemented the
antecedent parts of these rules on memristor crossbars through the
newly introduced concept, {\it i.e.} fuzzy minterms. Then, this
fuzzy XOR function is applied to consecutive pixels of the input
grayscale image to extract edges from it. Simulation result showed
that our fuzzy edge detector can effectively extract edges even in
noisy environment. It also has this advantage that it can extract
edges of any given image all at once in real-time. Finally, it is
worth to mention that although in this study we concentrated on the
application of edge detection, it is obvious that our proposed
structure can be used for the implementation of other fuzzy
inference systems which use fuzzy rule base to make inference.

\end{document}